\documentclass[runningheads]{llncs}
\pdfoutput=1

\usepackage{graphicx}
\usepackage{comment}
\usepackage{amsmath,amssymb} 
\usepackage{color}

\usepackage{times}
\usepackage{epsfig}
\usepackage{diagbox}
\usepackage{adjustbox}
\usepackage{subcaption}
\usepackage{makecell}

\DeclareMathOperator*{\argmin}{arg\,min}
\usepackage{ruler}
\usepackage[width=122mm,left=12mm,paperwidth=146mm,height=193mm,top=12mm,paperheight=217mm]{geometry}

\begin{document}
\pagestyle{headings}
\mainmatter

\def\ECCVSubNumber{3338}  

\title{PUGeo-Net: A Geometry-centric Network for 3D Point Cloud Upsampling}


\titlerunning{PUGeo-Net}
%
\author{Yue Qian\inst{1} \and
Junhui Hou\inst{1} \and
Sam Kwong\inst{1} \and Ying He \inst{2}}
\authorrunning{Y. Qian et al.}
%
\institute{Department of Computer Science, City University of Hong Kong \\ \email{\{yueqian4-c,jh.hou, cssamk\}@cityu.edu.hk}\\ \and
School of Computer Science and Engineering, Nanyang Technological University \email{yhe@ntu.edu.sg}}

\maketitle

\begin{abstract}
This paper addresses the problem of generating uniform dense point clouds to describe the underlying geometric structures from given sparse point clouds. Due to the irregular and unordered nature, point cloud densification as a generative task is challenging. To tackle the challenge, we propose a novel deep neural network based method, called PUGeo-Net, that incorporates discrete differential geometry into deep learning elegantly, making it fundamentally different from the existing deep learning methods that are largely motivated by the image super-resolution techniques and generate new points in the abstract feature space. Specifically, our method learns the first and second fundamental forms, which are able to fully represent the local geometry unique up to rigid motion. We encode the first fundamental form in a $3\times 3$ linear transformation matrix $\bf T$ for each input point. Such a matrix approximates the augmented Jacobian matrix of a local parameterization that encodes the intrinsic information and builds a one-to-one correspondence between the 2D parametric domain and the 3D tangent plane, so that we can lift the adaptively distributed 2D samples (which are also learned from data) to 3D space. After that, we use the learned second fundamental form to compute a normal displacement for each generated sample and project it to the curved surface. As a by-product, PUGeo-Net can compute normals for the original and generated points, which is highly desired the surface reconstruction algorithms. 
We interpret PUGeo-Net using the local theory of surfaces in differential geometry, which is also confirmed by quantitative verification.
We evaluate PUGeo-Net on a wide range of 3D models with sharp features and rich geometric details and observe that PUGeo-Net, the first neural network that can jointly generate vertex coordinates and normals, consistently outperforms the state-of-the-art in terms of accuracy and efficiency for upsampling factor $4\sim 16$. In addition, PUGeo-Net can handle noisy and non-uniformly distributed inputs well, validating its robustness.
\end{abstract}

\vspace{-1cm}
\section{Introduction}
\vspace{-0.1in}
Three-dimensional (3D) point clouds, as the raw representation of 3D data, are used in a wide range of applications, such as 3D immersive telepresence \cite{orts2016holoportation}, 
3D city reconstruction \cite{lafarge2012creating}, \cite{musialski2013survey}, cultural heritage reconstruction \cite{xu2014tridimensional}, \cite{bolognesi2015testing}, geophysical information systems \cite{paine2016shoreline}, \cite{nie2016estimating}, autonomous driving \cite{chen2017multi}, \cite{li20173d},  simultaneous localization and mapping \cite{fioraio2011realtime}, \cite{cole2006using}, and  virtual/augmented reality \cite{held20123d}, \cite{santana2017multimodal}, just to name a few. Though recent years have witnessed great progress on the 3D sensing technology \cite{hakala2012full}, \cite{kimoto2014development}, it is still costly and time-consuming to obtain dense and highly detailed point clouds, which are beneficial to the subsequent applications. 
Therefore, amendment is required to speed up the deployment of such data modality. 
In this paper, instead
of relying on the development of hardware, we are interested in the problem of computational based point cloud upsampling: given a sparse, low-resolution point cloud, generate a uniform and dense point cloud with a typical computational method to faithfully represent the underlying surface. Since the problem is the 3D counterpart of image super-resolution \cite{lai2017deep}, \cite{zhang2018residual}, a typical idea is to borrow the powerful techniques from the image processing community. However, due to the unordered and irregular nature of point clouds, such an extension is far from trivial, especially when the underlying surface has complex geometry.

The existing methods for point cloud upsampling can be roughly classified into two categories: optimization-based methods and deep learning based methods.
The optimization methods \cite{alexa2003computing}, \cite{lipman2007parameterization}, \cite{huang2009consolidation}, \cite{preiner2014continuous}, \cite{huang2013edge} usually fit local geometry and work well for smooth surfaces with less features. However, these methods struggle with multi-scale structure preservation. The deep learning methods can effectively learn structures from data. Representative methods are PU-Net \cite{yu2018pu}, EC-Net \cite{yu2018ec} and MPU \cite{yifan2019patch}. PU-Net extracts multi-scale features using point cloud convolution \cite{qi2017pointnet++} and then expands the features by replication. With additional edge and surface annotations, EC-Net improves PU-Net by restoring sharp features. Inspired by image super-resolution, MPU upsamples points in a progressive manner, where each step focuses on a different level of detail. PU-Net, EC-Net and MPU operate on patch level, therefore, they can handle high-resolution point sets. Though the deep learning methods produce better results than the optimization based methods, they are heavily motivated by the techniques in the image domain and takes little consideration of the geometries of the input shape. As a result, various artifacts can be observed in their results. It is also worth noting that all the existing deep learning methods generate points only, none of them is able to estimate the normals of the original and generated points.

\begin{figure*}[t]
\centering
\includegraphics[width=1\textwidth]{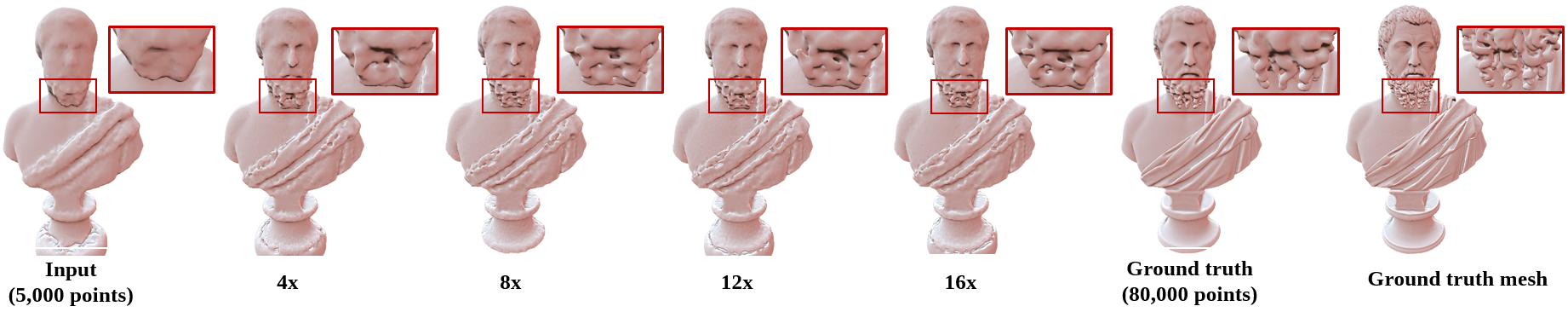}
\vspace{-0.3in}
\caption{\footnotesize{Illustration of various sampling factors of the Retheur Statue model with 5,000 points. Due to the low-resolution input, the details, such as cloth wrinkles and facial features, are missing. PUGeo-Net can effectively generate up to $16\times$ points to fill in the missing part. See also the accompanying video and results.} 
}
\label{fig:multirate} \vspace{-0.35in}
\end{figure*}

In this paper, we propose a novel network, called PUGeo-Net, to overcome the limitations in the existing deep learning methods. Our method learns a local parameterization for each point and its normal direction. In contrast to the existing neural network based methods that generate new points in the abstract feature space and map the samples to the surface using decoder, PUGeo-Net performs the sampling operations in a pure geometric way. Specifically, it first generates the samples in the 2D parametric domain and then lifts them to 3D space using a linear transformation. Finally, it projects the points on the tangent plane onto the curved surface by computing a normal displacement for each generated point via the learned second fundamental form. Through extensive evaluation on commonly used as well as new metrics, we show that PUGeo-Net consistently outperforms the state-of-the-art in terms of accuracy and efficiency for upsampling factors $4\sim 16\times$. It is also worth noting that PUGeo-Net is the first neural network that can generate dense point clouds with accurate normals, which are highly desired by the existing surface reconstruction algorithms. We demonstrate the efficacy of PUGeo-Net on both CAD models with sharp features and scanned models with rich geometric details and complex topologies.
Fig.~\ref{fig:multirate} demonstrates the effectiveness of PUGeo-Net on the \textit{Retheur} Statue model.

The main contributions of this paper are summarized as follows. \vspace{-0.3cm}
\begin{enumerate}
\item We propose PUGeo-Net, a novel geometric-centric neural network, which carries out a sequence of geometric operations, such as computing the first-order approximation of local parameterization, adaptive sampling in the parametric domain, lifting the samples to the tangent plane, and projection to the curved surface. 
\item PUGeo-Net is the first upsampling network that can jointly generate coordinates and normals for the densified point clouds. The normals benefit many downstream applications, such as surface reconstruction and shape analysis.
\item We interpret PUGeo-Net using the local theory of surfaces in differential geometry. Quantitative verification confirms our interpretation. 
\item We evaluate PUGeo-Net on both synthetic and real-world models and show that PUGeo-Net significantly outperforms the state-of-the-art methods in terms of accuracy and efficiency for all upsampling factors.
\item  PUGeo-Net can handle noisy and non-uniformly distributed point clouds as well as the real scanned data by the LiDAR sensor very well, validating its robustness and practicality.
\end{enumerate}

\vspace{-0.25in}
\section{Related Work}
\vspace{-0.15in}

{\bf Optimization based methods.} 
Alexa \textit{et al.} \cite{alexa2003computing} interpolated points of Voronoi diagram, which is computed in the local tangent space. Lipman \textit{et al.} developed a method based on locally optimal projection operator (LOP) \cite{lipman2007parameterization}. It is a parametrization-free method for point resampling and surface reconstruction. Subsequently, the improved weighted LOP and continuous LOP were developed by Huang \textit{et al.} \cite{huang2009consolidation} and  Preiner \textit{et al.} \cite{preiner2014continuous}  respectively. These methods assume that points are sampling from smooth surfaces, which degrades upsampling quality towards sharp edges and corners. Huang \textit{et al.} \cite{huang2013edge} presented an edge-aware (EAR) approach which can effectively preserve the sharp features. With given normal information, EAR algorithm first resamples points away from edges, then progressively upsamples points to approach the edge singularities. However, the performance of EAR heavily depends on the given normal information and parameter tuning. In conclusion, point cloud upsampling methods based on geometric priors either assume insufficient hypotheses or require additional attributes.

{\bf Deep learning based methods.} The deep learning based upsampling methods first extract point-wise feature via point clouds CNN. The lack of point order and regular structure impede the extension of powerful CNN to point clouds. Instead of converting point clouds to other data representations like volumetric grids \cite{maturana2015voxnet}, \cite{riegler2017octnet}, \cite{wu20153d} or graphs \cite{landrieu2018large}, \cite{te2018rgcnn}, recently the point-wise 3D CNN \cite{qi2017pointnet}, \cite{wang2018dynamic}, \cite{qi2017pointnet++}, \cite{komarichev2019cnn}, \cite{li2018pointcnn} successfully achieved state-of-the-art performance for various tasks. 
Yu \textit{et al.} pioneered PU-Net\cite{yu2018pu}, the first deep learning algorithm for point cloud upsampling. It adopts PointNet++ \cite{qi2017pointnet++} to extract point features and expands features by multi-branch MLPs. It optimizes a joint reconstruction and repulsion loss function to generate point clouds with uniform density. PU-Net surpasses the previous optimization based approaches for point cloud upsampling.
However, as it does not consider the spatial relations among the points, there is no guarantee that the generated samples are uniform. The follow-up work, EC-Net \cite{yu2018ec}, adopts a joint loss of 
point-to-edge distance, which can effectively preserve sharp edges. EC-Net requires labelling the training data with annotated edge and surface information, which is tedious to obtain.
Wang \textit{et al.} \cite{yifan2019patch} proposed a patch-based progressive upsampling method (MPU). Their method can successfully apply to large upsampling factor, say 16$\times$. Inspired by the image super-resolution techniques, they trained a cascade of upsampling networks to progressively upsample to the desired factor, with the subnet only deals with 2$\times$ case. MPU replicates the point-wise features and separates them by appending a 1D code $\{-1, 1\}$, which does not consider the local geometry. MPU requires a careful step-by-step training, which is not flexible and fails to gain a large upsampling factor model directly.
Since each subnet upsizes the model by a factor 2, MPU only works for upsampling factor which is a power of 2. PUGeo-Net distinguishes itself from the other deep learning method from its geometry-centric nature. See Sec.~\ref{sec:exp} for quantitative comparisons and detailed discussions. Recently, Li \textit{et al.} \cite{li2019pu} proposed PU-GAN  which introduces an adversarial framework to train the upsampling generator. Again, PU-GAN fails to examine the geometry properties of point clouds. Their ablation studies also verify the performance improvement mainly comes from the introducing of the discriminator.

\vspace{-0.15in}
\section{Proposed Method}
\vspace{-0.3cm}
\subsection{Motivation \& Overview}\label{subsec:motivation} \vspace{-0.3cm}
Given a sparse point cloud $\mathcal{X}=\{\mathbf{x}_i\in\mathbb{R}^{3\times 1}\}_{i=1}^M$ with $M$ points and the user-specified upsampling factor $R$, we aim to generate a dense, uniformly distributed point cloud $\mathcal{X}_R=\{\mathbf{x}_i^{r}\in\mathbb{R}^{3\times 1}\}_{i, r=1}^{M, R}$, which contains more geometric details and can approximate the underlying surface well. Similar to other patch-based approaches, we first partition the input sparse point cloud into patches via the farthest point sampling algorithm, each of which has $N$ points, and PUGeo-Net processes the patches separately. 

As mentioned above, the existing deep learning based methods are heavily built upon the techniques in 2D image domain, which generate new samples by replicating feature vectors in the abstract feature space, and thus the performance is limited. Moreover, due to little consideration of shape geometry, none of them can compute normals, which play a key role in surface reconstruction. In contrast, our method is motivated by parameterization-based surface resampling, consisting of 3 steps: first it parameterizes a 3D surface $S$ to a 2D domain, then it samples in the parametric domain and finally maps the 2D samples to the 3D surface. It is known that parameterization techniques depend heavily on the topology of the surface. There are two types of parameterization, namely local parameterization and global parameterization. The former deals with a topological disk (i.e., a genus-0 surface with 1 boundary)~\cite{MIPS}. The latter works on surfaces of arbitrary topology by computing canonical homology basis, through which the surface is cutting into a topological disk, which is then mapped to a 2D domain~\cite{GuY03}. Global constraints are required in order to ensure the parameters are continuous across the cuts~\cite{Campen}.
\begin{figure}[h!]
\centering
\vspace{-0.15in}
\includegraphics[width=0.9\textwidth]{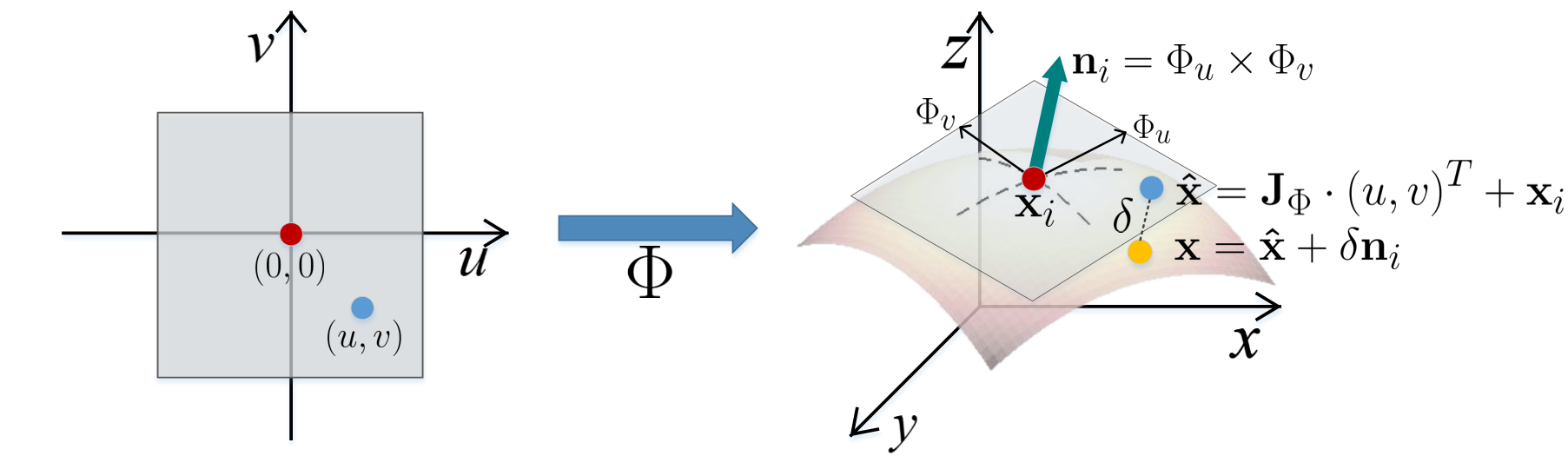}
\vspace{-0.2in}
\caption{\footnotesize{Surface parameterization and local shape approximation. The local neighborhood of $\mathbf{x}_i$ is parameterized to a 2D rectangular domain via a differentiable map $\mathbf{\Phi}:\mathbb{R}^2\rightarrow\mathbb{R}^3$. The Jacobian matrix $\mathbf{J}_\Phi(0,0)$ provides the best linear approximation of $\mathbf{\Phi}$ at $\mathbf{x}_i$, which maps $(u,v)$ to a point $\mathbf{\hat{x}}$ on the tangent plane of $\mathbf{x}_i$. Furthermore, using the principal curvatures of $\mathbf{x}_i$, we can reconstruct the local geometry of $\mathbf{x}_i$ in the second-order accuracy.}}
\label{fig:parametrize}\vspace{-0.3in}
\end{figure}

In our paper, the input is a point cloud sampled from a 3D surface of arbitrary geometry and topology. The Fundamental Theorem of the Local Theory of Surfaces states that the local neighborhood of a point on a regular surface can be completely determined by the first and second fundamental forms, unique up to rigid motion (see \cite{decamero}, Chapter 4). Therefore, instead of computing and learning a \textit{global} parameterization which is expensive, our key idea is to learn a \textit{local} parameterization for each point.

Let us parameterize a local neighborhood of point $\mathbf{x}_i$ to a 2D domain via a differential map $\mathbf{\Phi}:\mathbb{R}^2\rightarrow\mathbb{R}^3$ so that $\mathbf{\Phi}(0,0)=\mathbf{x}_i$ (see Fig.~\ref{fig:parametrize}). The Jacobian matrix $\mathbf{J}_{\Phi}=\left[\mathbf{\Phi}_u,\mathbf{\Phi}_v\right]$ provides the best first-order approximation of the map $\mathbf{\Phi}$: 
$\mathbf{\Phi}(u,v)=\mathbf{\Phi}(0,0)+\left[\mathbf{\Phi}_u,\mathbf{\Phi}_v\right]\cdot(u,v)^\textsf{T} + O(u^2+v^2)$,  
where $\mathbf{\Phi}_u$ and $\mathbf{\Phi}_v$ are the tangent vectors, which define the first fundamental form. The normal of point $\mathbf{x}_i$ can be computed by the cross product $\mathbf{n}_i=\mathbf{\Phi}_u(0,0)\times\mathbf{\Phi}_v(0,0)$.

It is easy to verify that the point $\mathbf{\hat{x}}\triangleq\mathbf{x}_i+\mathbf{J}_{\Phi}\cdot(u,v)^\textsf{T}$ is on the tangent plane of $\mathbf{x}_i$, 
since $(\mathbf{\hat{x}}-\mathbf{x}_i)\cdot\mathbf{n}_i=0$. 
In our method, we use the augmented Jacobian matrix 
$\mathbf{T}=[\mathbf{\Phi}_{u}, \mathbf{\Phi}_v, \mathbf{\Phi}_u\times\mathbf{\Phi}_v] $
to compute the normal 
$\mathbf{n}_i=\mathbf{T}\cdot(0,0,1)^{\textsf{T}}$ 
and the point $
\mathbf{\hat{x}}=\mathbf{x}_i+\mathbf{T}\cdot(u,v,0)^{\textsf{T}}$. Matrix $\bf T$ is of full rank if the surface is regular at $\mathbf{x}_i$. 
Furthermore, the distance between $\mathbf{x}$ and $\mathbf{\hat{x}}$ is
$\|\mathbf{x}-\mathbf{\hat{x}}\|=\frac{\kappa_1 u^2+\kappa_2 v^2}{2}+O(u^3+v^3)$, where $\kappa_1$ and $\kappa_2$ are the principal curvatures at $\mathbf{\Phi}(0,0)$, which are the eigenvalues of the second fundamental form.


\begin{figure*}[t]
     \centering
     \begin{subfigure}[b]{1\textwidth}
         \centering
         \includegraphics[width=\textwidth]{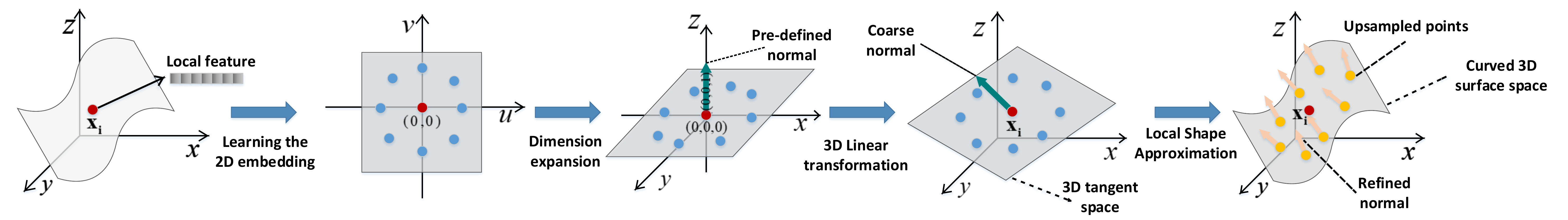}
     \end{subfigure}
     \hfill
     \begin{subfigure}[b]{1\textwidth}
         \centering
         \includegraphics[width=\textwidth]{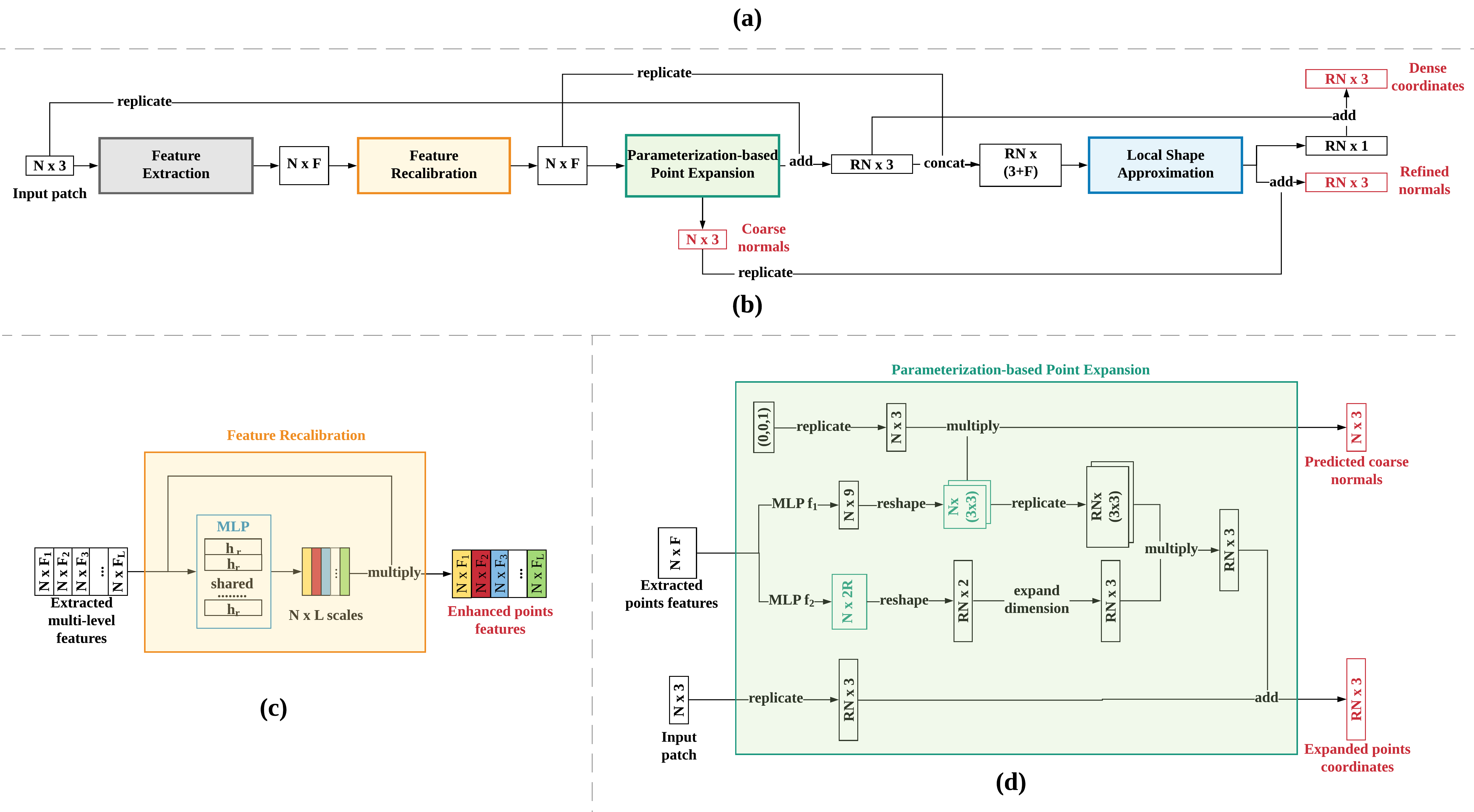}
     \end{subfigure} \vspace{-0.3in}
       \caption{\footnotesize{Overview. The top row illustrates the stages of PUGeo-Net: learning local parameterization, point expansion and vertex coordinates and normals refinement. The middle row shows the end-to-end network structure of PUGeo-Net. The output is colored in red. The bottom row shows the details of two core modules: feature recalibration and point expansion.}}     
        \label{fig:flow} \vspace{-0.3in}
\end{figure*}
As shown in Fig. \ref{fig:flow}(a), given an input sparse 3D point cloud, PUGeo-Net proceeds as follows: 
it first generates new samples $\{(u_i^r,v_i^r)\}_{r=1}^R$ in the 2D parametric domain. Then it computes the normal $\mathbf{n}_i=\mathbf{T}_i \cdot(0,0,1)^{\textsf{T}}$. After that, it maps each generated 2D sample $(u_i,v_i)$ to the tangent plane of $\mathbf{x}_i$ by  $\mathbf{\widehat{x}}_i^r=\mathbf{T}_i\cdot(u_i^r,v_i^r,0)^\textsf{T}+\mathbf{x}_i$. Finally, it projects $\mathbf{\hat{x}}_i^r$ to the curved 3D surface by computing a displacement $\delta_i^r$ along the normal direction.  Fig. \ref{fig:flow}(b) illustrates the network architecture of PUGeo-Net,
which consists of hierarchical feature extraction and re-calibration (Sec. \ref{subsec:feature}), parameterization-based point expansion (Sec. \ref{subsec:expansion}) and local shape approximation (Sec. \ref{subsec:refine}). We adopt a joint loss function to guide the prediction of vertex coordinates and normals (Sec \ref{subsec:loss}). 
\vspace{-0.2in}

\subsection{Hierarchical Feature Learning and Recalibration}\label{subsec:feature}\vspace{-0.25cm}

To handle the rotation-invariant challenge of 3D point clouds, we adopt an STN-like mini-network~\cite{jaderberg2015spatial}, which computes a global 3D transformation matrix $\mathbf{A} \in \mathbb{R}^{3\times 3}$ applied to all points. After that, we apply DGCNN \cite{wang2018dynamic} - the state-of-the-art method for point cloud classification and segmentation - to extract hierarchical point-wise features, which are able to encode both local and global intrinsic geometry information of an input patch.   

The hierarchical feature learning module extracts features from low- to high-levels. Intuitively speaking, as the receptive fields increase, skip-connection \cite{ronneberger2015u}, \cite{he2016deep}, \cite{huang2017densely}, a  widely-used technique in 2D vision task for improving the feature quality and the convergence speed, can help preserve details in all levels. 
To this end, as illustrated in Fig. \ref{fig:flow}(c), instead of concatenating the obtained features directly, we perform feature re-calibration by a  self-gating unit \cite{hu2018squeeze}, \cite{zhang2019self} to enhance them, which is computationally efficient. 

Let  $\mathbf{c}_i^{l}\in\mathbb{R}^{F_l\times 1}$ be the extracted feature for point $\mathbf{x}_i$ at the $l$-th level ($l=1,\cdots,L$), where $F_l$ is the feature length.  
We first concatenate the features of all $L$ layers, i.e., $\mathbf{\widehat{c}}_i=\textsf{Concat}(\mathbf{c}_i^1, \cdots, \mathbf{c}_i^L) \in \mathbb{R}^{F}$,  where $F=\sum_{l=1}^L F_l$ and $\textsf{Concat} (\cdot)$ stands for the concatenation operator. The direct concatenate feature is passed to a small MLP $h_r (\cdot)$ to obtain the logits $\mathbf{a}_i=(a_i^1, a_i^2,...,a_i^L)$, i.e., \vspace{-0.3cm}
\begin{equation}
    \mathbf{a}_i = h_r(\mathbf{\hat{c}}_i), \vspace{-0.3cm}
\end{equation}
which are futher fed to a softmax layer to produce the recalibration weights  $\mathbf{w}_i=(w_i^1, w_i^2, \cdots, w_i^L)$ with \vspace{-0.3cm}
\begin{equation}
    w_i^l = e^{a_i^l}/\sum_{k=1}^L e^{a_i^k}. \vspace{-0.3cm}
\end{equation}
Finally, the recalibrated multi-scale features are represented as the weighted concatenation:\vspace{-0.2cm}
\begin{equation}
    \mathbf{c}_i=\textsf{Concat}(w_i^1\cdot\mathbf{c}_i^1, w_i^2\cdot\mathbf{c}_i^2, \cdots, \hat{a}_i^L\cdot \mathbf{c}_i^L).
\end{equation}

     \vspace{-0.25in}
\subsection{Parameterization-based Point Expansion}\label{subsec:expansion}\vspace{-0.25cm}
In this module, we expand the input spare point cloud $R$ times to generate a coarse dense point cloud as well as the corresponding coarse normals by regressing the obtained multi-scale features. Specifically, the expansion process is composed of two steps, i.e., learning an adaptive sampling in the 2D parametric domain and then projecting it onto the 3D tangent space by a learned linear transformation. 

\textbf{Adaptive sampling in the 2D parametric domain}. For each point $\mathbf{x}_i$, we apply an MLP $f_1(\cdot)$ to its local surface feature $\mathbf{c}_i$ to reconstruct the 2D coordinates $(u_i^r,v_i^r)$ of $R$ sampled points, i.e., \vspace{-0.3cm}
\begin{equation}
\{(u_i^r, v_i^r)| r=1,2,  \cdots,R\}=f_1(\mathbf{c}_i).\vspace{-0.3cm}
\end{equation}
With the aid of its local surface information encoded in $\mathbf{c}_i$, it is expected that the self-adjusted 2D parametric domain maximizes the uniformity over the underlying surface.  

\textit{Remark}. Our sampling strategy is fundamentally different from the existing deep learning methods. PU-Net generates new samples by replicating features in the feature space, and feed the duplicated features into independent multi-branch MLPs. 
It adopts an additional repulsion loss to regularize uniformity of the generated points. MPU also replicates features in the feature space. It appends additional code $+1$ and $-1$ to the duplicated feature copies in order to separate them. Neither PU-Net nor MPU considers the spatial correlation among the generated points. In contrast, our method expands points in the 2D parametric domain and then lifts them to the tangent plane, hereby in a more geometric-centric manner. By viewing the problem in the mesh parametrization sense, we can also regard appending 1D code in MPU as a \textit{predefined} 1D parametric domain. Moreover, the predefined 2D regular grid is also adopted by other deep learning based methods for processing 3D point clouds, e.g., FoldingNet \cite{yang2018foldingnet}, PPF-FoldNet \cite{deng2018ppfnet} and PCN \cite{yuan2018pcn}. 
Although the predefined 2D grid is regularly distributed in 2D domain, it does not imply the transformed points are uniformly distributed on the underlying 3D surface.

\textbf{Prediction of the linear transformation}.
For each point $\mathbf{x}_i$, we also predict a linear transformation matrix $\mathbf{T}_i \in \mathbb{R}^{3\times 3}$ from the local surface feature $\mathbf{c}_i$, i.e., 
\vspace{-0.3cm}
\begin{equation}
    \mathbf{T}_i =f_2(\mathbf{c}_i),  \vspace{-0.3cm}
\end{equation}
where $f_2(\cdot)$ denotes an MLP. Multiplying $\mathbf{T}_i$ to the previously learned 2D samples $\{(u_i^r,v_i^r)\}_{r=1}^R$ lifts the points to the tangent plane of $\mathbf{x}_i$ \vspace{-0.2cm} 
\begin{equation}
    \widehat{\mathbf{x}}_i^r=(\widehat{x}_i^r,\widehat{y}_i^r,\widehat{z}_i^r)^\textsf{T}=\mathbf{T}_i\cdot(u_i^r,v_i^r,0)^\textsf{T}+\mathbf{x}_i. \vspace{-0.2cm}
\end{equation}

\textbf{Prediction of the coarse normal}. As aforementioned, normals of points play an key role in surface reconstruction. In this module, we first estimate a coarse normal, i.e., the normal $\mathbf{n}_i\in\mathbb{R}^{3\times 1}$ of the \textit{tangent plane} of each input point, which are shared by all points on it. Specifically, we multiply the 
linear transformation matrix $\mathbf{T}_i$ to the predefined normal $(0,0,1)$ which is perpendicular to the 2D parametric domain:  \vspace{-0.3cm}
\begin{equation}
    \mathbf{n}_i = \mathbf{T}_i\cdot (0,0,1)^\textsf{T}.
\end{equation}

     \vspace{-0.3in}

\subsection{Updating Samples via Local Shape Approximation}\label{subsec:refine}
\vspace{-0.1in}

Since the samples $\widehat{\mathcal{X}}_R=\{\widehat{\mathbf{x}}_i^r\}_{i, r=1}^{M, R}$ are on the tangent plane, we need to warp them to the curved surface and update their normals. Specifically, we move each sample $\mathbf{\hat{x}}_i^r$ along the normal $\mathbf{n}_i$ with a distance $\delta_i^r=\frac{\kappa_1 (u_{i}^{r})^2+\kappa_2 (v_{i}^{r})^2}{2}$. As mentioned in Sec.~\ref{subsec:motivation}, this distance provides the second-order approximation of the local geometry of $\mathbf{x}_i$. 
We compute the distance $\delta_i^r$ by regressing
the point-wise features concatenated with their coarse coordinates, i.e., 
\vspace{-0.25cm}
\begin{equation}
\centering
\delta_i^r = f_3(\textsf{Concat}(\widehat{\mathbf{x}}_i^r, \mathbf{c}_i)),\vspace{-0.15cm} 
\end{equation}
where $f_3(\cdot)$ is for the process of an MLP. Then we compute the sample coordinates as  \vspace{-0.25cm}
\begin{equation}
\centering
    \mathbf{x}_i^r=(x_i^r,y_i^r,z_i^r)^\textsf{T} = 
\widehat{\mathbf{x}}_i^r + \mathbf{T}_i \cdot (0,0,\delta_i^r)^\textsf{T}. \vspace{-0.25cm} 
\end{equation}

We update the normals in a similar fashion: a normal offset $\Delta \mathbf{n}_i^r\in\mathbb{R}^{3\times 1}$ for point $\mathbf{x}_i^r$ is regressed as 
\vspace{-0.25cm}
\begin{equation}
 \centering
 \Delta \mathbf{n}_i^r = f_4\left(\textsf{Concat}( \widehat{\mathbf{x}}_i^r, \mathbf{c}_i)\right), \vspace{-0.25cm}
\end{equation}
which is further added to the corresponding coarse normal, leading to 
\vspace{-0.25cm}
\begin{equation}
    \mathbf{n}_i^r =\Delta \mathbf{n}_i^r + \mathbf{n}_i, \vspace{-0.25cm}
\end{equation}
where $f_4(\cdot)$ is the process of an MLP.
     \vspace{-0.15in}

\subsection{Joint Loss Optimization}\label{subsec:loss}\vspace{-0.25cm}
As PUGeo-Net aims to deal with the regression of both coordinates and unoriented normals of points, we design a joint loss to train it end-to-end. 
Specifically, let $\mathcal{Y}_R=\{\mathbf{y}_k\}_{k=1}^{RM}$ with $RM$ points be the groundtruth of $\mathcal{X}_R$. 
During training, we adopt the Chamfer distance (CD) to measure the coordinate error between the $\mathcal{X}_R$ and $\mathcal{Y}_R$, i.e., \vspace{-0.6cm} 
 
\begin{align}
 &L_{CD}=  
 \frac{1}{RM}\left(\sum_{\mathbf{x}_i^r\in \mathcal{X}_R}||\mathbf{x}_i^r-\phi(\mathbf{x}_i^r)||_2+\sum_{\mathbf{y}_k\in \mathcal{Y}_R}||\mathbf{y}_k-\psi(\mathbf{y}_k)||_2\right),\nonumber \vspace{-1cm}
\end{align} 
where $\phi(\mathbf{x}_i^r)=\argmin_{\mathbf{y}_k\in \mathcal{Y}_R}\|\mathbf{x}_i^r-\mathbf{y}_k\|_2, \psi(\mathbf{y}_k)=\argmin_{\mathbf{x}_i^r\in \mathcal{X}_R}\|\mathbf{x}_i^r-\mathbf{y}_k\|_2$, and $\|\cdot\|_2$ is the $\ell_2$ norm of a vector.

For the normal part, denote $\widetilde{\mathcal{N}}=\{\widetilde{\mathbf{n}}_i\}_{i=1}^M$ and $\overline{\mathcal{N}}_R=\{\overline{\mathbf{n}}_k\}_{k=1}^{RM}$ the ground truth of the coarse normal $\mathcal{N}$ and the accurate normal $\mathcal{N}_R$, respectively. During training, we consider the  errors between  $\mathcal{N}$ and $\widetilde{\mathcal{N}}$  and between $\mathcal{N}_R$ and $\overline{\mathcal{N}}_R$ simultaneously, i.e., \vspace{-0.4cm}
\begin{equation}
        L_{coarse}(\mathcal{N}, \widetilde{\mathcal{N}}) = \sum_{i=1}^M L(\mathbf{n}_i, \widetilde{\mathbf{n}}_i),~ 
L_{refined}(\mathcal{N}_R, \overline{\mathcal{N}}_R) = \sum_{i=1}^M\sum_{r=1}^R L(\mathbf{n}_i^r, \overline{\mathbf{n}}_{\phi(\mathbf{x}_i^r)}), \vspace{-0.3cm} 
\end{equation}
where $L(\mathbf{n}_i, \tilde{\mathbf{n}}_i) = \max\left\{\|\mathbf{n}_i- \tilde{\mathbf{n}}_i\|_2^2, \|\mathbf{n}_i+ \tilde{\mathbf{n}}_i\|_2^2\right\}$ measures the unoriented difference between two normals, and $\phi(\cdot)$ is used to build the unknown correspondence between $\mathcal{N}_R$ and $\overline{\mathcal{N}}_R$. 
Finally, the joint loss function is written as \vspace{-0.25cm}
\begin{equation}
    L_{total} = \alpha L_{CD}+\beta L_{coarse}+\gamma L_{refined}, \vspace{-0.25cm}
    \label{eqn:totalloss}
\end{equation}
where $\alpha$, $\beta$, and $\gamma$ are three positive parameters.
It is worth noting that our method does not require repulsion loss which is required by PU-Net and EC-Net, since the module for learning the parametric domain is capable of densifying point clouds with uniform distribution.

     \vspace{-0.2in}

\section{Experimental Results}
\label{sec:exp}
 \vspace{-0.35cm}
\subsection{Experiment Settings}
     \vspace{-0.1in}
\textbf{Datasets}. Following previous works, we selected 90 high-resolution 3D mesh models from Sketchfab \cite{sketchfab} to construct the training dataset and 13 for the testing dataset.  
Specifically, given the 3D meshes, we employed the Poisson disk sampling~\cite{corsini2012efficient} to generate $\mathcal{X}$, $\mathcal{Y}_R$, $\widetilde{\mathcal{N}}$, and $\overline{\mathcal{N}}$ with $M=5000$ and $R=4, 8, 12$ and $16$. 
A point cloud was randomly cropped into patches each of $N=256$ points.   
To \textit{fairly} compare different methods, we adopted identical data augmentations settings, including random scaling, rotation and point perturbation. During the testing process, clean test data were used. Also notice that the normals of sparse inputs are not needed during testing.

\textbf{Implementation details}. 
We empirically set the values of the three parameters $\alpha$, $\beta$, and $\gamma$ in the joint loss function to $100$, $1$, and $1$, respectively. We used the Adam algorithm with the learning rate equal to 0.001. We trained the network with the mini-batch of size 8 for 800 epochs via the TensorFlow platform. \textit{The code will be publicly available later}.

\textbf{Evaluation metrics}. To quantitatively evaluate the performance of different methods, we considered four commonly-used evaluation metrics, i.e., Chamfer distance (CD), Hausdorff distance (HD), point-to-surface distance (P2F), and Jensen-Shannon divergence (JSD). For these four metrics, the lower, the better. For all methods under comparison, we applied the metrics on the whole shape. 

We also propose a new approach to quantitatively measure the quality of the generated point clouds. Instead of conducting the comparison between the generated point clouds and the corresponding groundtruth ones directly, we first performed surface reconstruction \cite{kazhdan2013screened}. For the methods that cannot generate normals 
principal component analysis (PCA) was adopted to predict normals. Then we densely sampled $200,000$ points from reconstructed surface. CD, HD and JSD between the densely sampled points from reconstructed surface and the groundtruth mesh were finally computed for measuring the surface reconstruction quality. Such new measurements are denoted as CD$^\#$, HD$^\#$ and JSD$^\#$.

\vspace{-0.4cm}
\subsection{Comparison with State-of-the-art Methods}\vspace{-1.2cm}

\begin{table*}
\centering
\caption{\footnotesize{Results of quantitative comparisons. The models were scaled uniformly in a unit cube, so the error metrics are unitless. Here, the values are the average of 13 testing models. See the \textit{Supplementary Material} for the results of each model.}
}
\scriptsize
\begin{tabular}{c|c||c |c c c c c|c c c}\Xhline{5\arrayrulewidth}
$R$ & Method &\makebox[3em]{Network} & \makebox[3em]{CD } & \makebox[3em]{HD }  &\makebox[3em]{JSD } & \makebox[3.5em]{P2F mean} & \makebox[3.5em]{P2F std} &\makebox[4em]{CD$^\#$} & \makebox[3em]{HD$^\#$} & \makebox[3em]{JSD$^\#$} \\
       &  &\makebox[3em]{size} & \makebox[3em]{$(10^{-2})$} & \makebox[3em]{$(10^{-2})$}  &\makebox[3em]{$(10^{-2})$} & \makebox[3em]{$(10^{-3})$} & \makebox[3em]{$(10^{-3})$} &\makebox[3em]{$(10^{-2})$} & \makebox[3em]{$(10^{-2})$} & \makebox[3em]{$(10^{-2})$} \\
\Xhline{2\arrayrulewidth}
4$\times$ & EAR \cite{huang2013edge} & - & 0.919 & 5.414 & 4.047 & 3.672 & 5.592 & 1.022 & 6.753 & 7.445 \\
& PU-Net \cite{yu2018pu} & 10.1 MB & 0.658 & 1.003 & 0.950 & 1.532 & 1.215 & 0.648 & 5.850 & 4.264 \\
& MPU \cite{yifan2019patch}& 92.5 MB & 0.573 & 1.073 & 0.614 & 0.808 & 0.809 & 0.647 & 5.493 & 4.259 \\
&PUGeo-Net & 26.6 MB & \bf{0.558} & \bf{0.934} & \bf{0.444} & \bf{0.617} & \bf{0.714} & \bf{0.639} & \bf{5.471} & \bf{3.928} \\\Xhline{2\arrayrulewidth}
8$\times$ & EAR \cite{huang2013edge} & - & - & - & - & - & - & - & - & -\\
&PU-Net \cite{yu2018pu} & 14.9 MB & 0.549 & 1.314 & 1.087 & 1.822 & 1.427 & 0.594 & 5.770 & 3.847 \\
&MPU \cite{yifan2019patch}& 92.5 MB & 0.447 & 1.222 & 0.511 & 0.956 & 0.972 & 0.593 & 5.723 & 3.754 \\
& PUGeo-Net & 26.6 MB & \bf{0.419} & \bf{0.998} & \bf{0.354} & \bf{0.647} & \bf{0.752} & \bf{0.549} & \bf{5.232} & \bf{3.465} \\\Xhline{2\arrayrulewidth}
12$\times$ & EAR \cite{huang2013edge} & - & - & - & - & - & - & - & - & -\\
& PU-Net \cite{yu2018pu} & 19.7 MB & 0.434 & 0.960 & 0.663 & 1.298 & 1.139 & 0.573 & 6.056 & 3.811 \\
& MPU \cite{yifan2019patch}& - & - & - & - & - & - & - & - & -\\
& PUGeo-Net & 26.7 MB & \bf{0.362} & \bf{0.978} & \bf{0.325} & \bf{0.663} & \bf{0.744}  & \bf{0.533} & \bf{5.255} & \bf{3.322} \\\Xhline{2\arrayrulewidth}
16$\times$ & EAR \cite{huang2013edge} & - & - & - & - & - & - & - & - & -\\
& PU-Net \cite{yu2018pu} & 24.5 MB & 0.482 & 1.457 & 1.165 & 2.092 & 1.659 & 0.588 & 6.330 & 3.744 \\
& MPU \cite{yifan2019patch} & 92.5 MB & 0.344 & 1.355 & 0.478 & 0.926 & 1.029 & 0.573 & 5.923 & 3.630 \\
& PUGeo-Net & 26.7 MB & \bf{0.323} & \bf{1.011} & \bf{0.357}& \bf{0.694} & \bf{0.808} & \bf{0.524} & \bf{5.267} & \bf{3.279} \\\Xhline{5\arrayrulewidth} 
\multicolumn{11}{l}{CD$^\#$, HD$^\#$, JSD$^\#$: these 3 metrics are used to measure the distance between dense point clouds}\\
\multicolumn{11}{l}{sampled from reconstructed surfaces and ground truth meshes.}
\end{tabular}
\label{table:compare} 
\vspace{-0.8cm}
\end{table*}

We compared PUGeo-Net with three methods, i.e., optimization based EAR \cite{huang2013edge}, and two state-of-the-art deep learning based methods, i.e., PU-Net \cite{yu2018pu} and MPU \cite{yifan2019patch}.
For fair comparisons, we retrained PU-Net and MPU with the same dataset as ours. Notice that EAR fails to process the task with $R$ greater than 4, due to the huge memory consumption,  and MPU can work only for tasks with $R$ in the powers of 2, due to its natural cascaded structure. \textit{Note that the primary EAR, PU-Net and MPU cannot predict normals}.

\textbf{Quantitative comparisons.} Table \ref{table:compare} shows the average result of 13 testing models, where we can observe that PUGeo-Net can achieve the best performance for \textit{all} upsample factors in terms of \textit{all} metrics. 
Moreover, the network size of PUGeo-Net is fixed and much smaller than that of MPU. Due to the deficiency of the independent multi-branch design, the network size of PU-Net grows linearly with the the upsample factor increasing, and is comparable to ours when $R=16$. 

\textbf{Visual comparisons.} The superiority of PUGeo-Net is also visually demonstrated.  We compared the reconstructed surfaces from the input sparse point clouds and the generated dense point clouds by different methods. Note that the surfaces were reconstructed via the same method as \cite{kazhdan2013screened}, in which the parameters ``depth" and ``minimum number of samples" were set as $9$ and $1$, respectively. For PU-Net and MPU which fail to predict normals, we adopted PCA normal estimation with the neighbours equal to $16$. Here we took the task with $R=16$ as an example. Some parts highlighted in red and blue boxes are zoomed in for a close look.       
     \vspace{-0.2in}

\begin{figure*}
     \centering
     \begin{subfigure}[b]{1\textwidth}
         \centering
         \includegraphics[width=\textwidth]{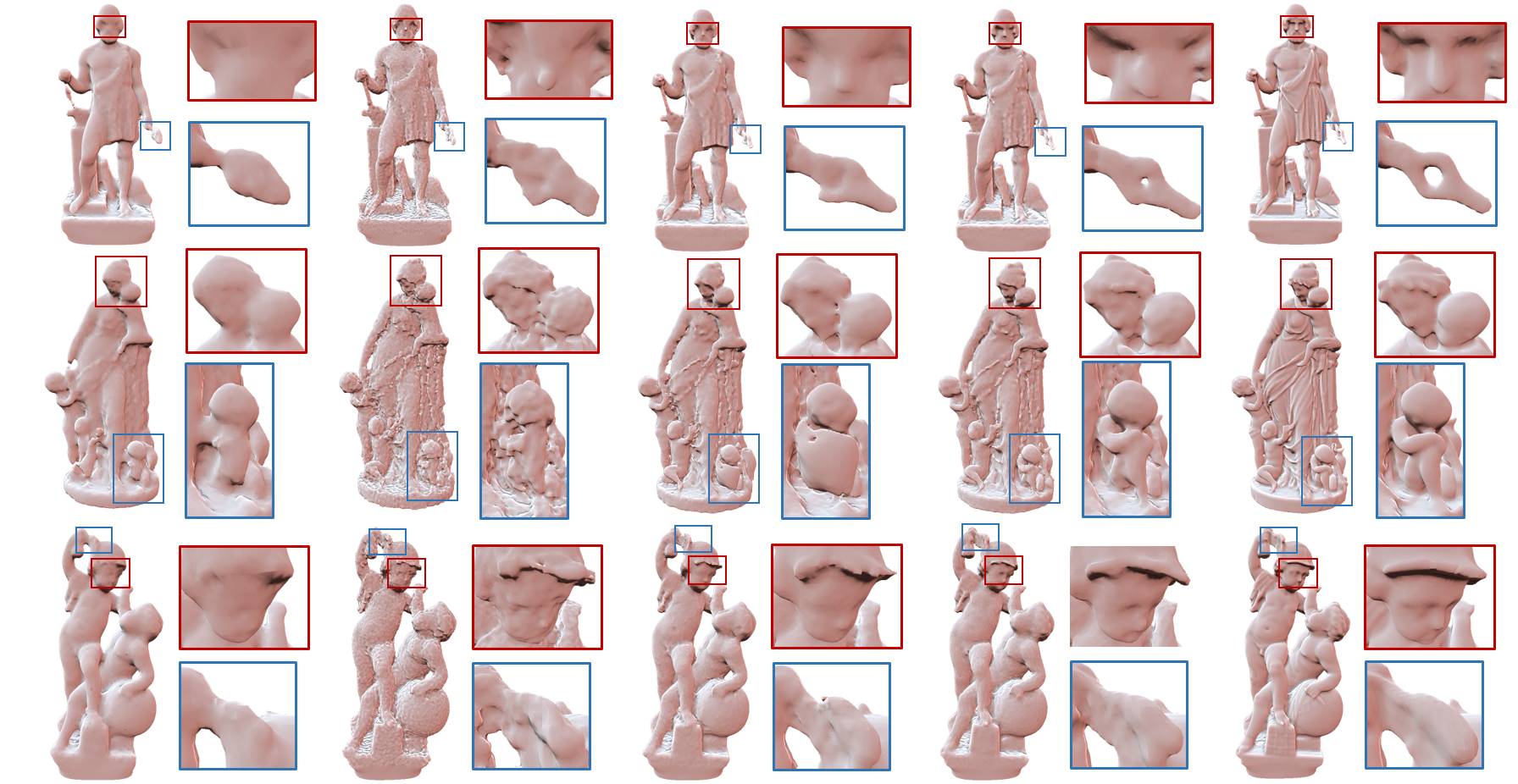}
     \end{subfigure}
     \hfill
     \begin{subfigure}[b]{1\textwidth}
         \centering
         \includegraphics[width=\textwidth]{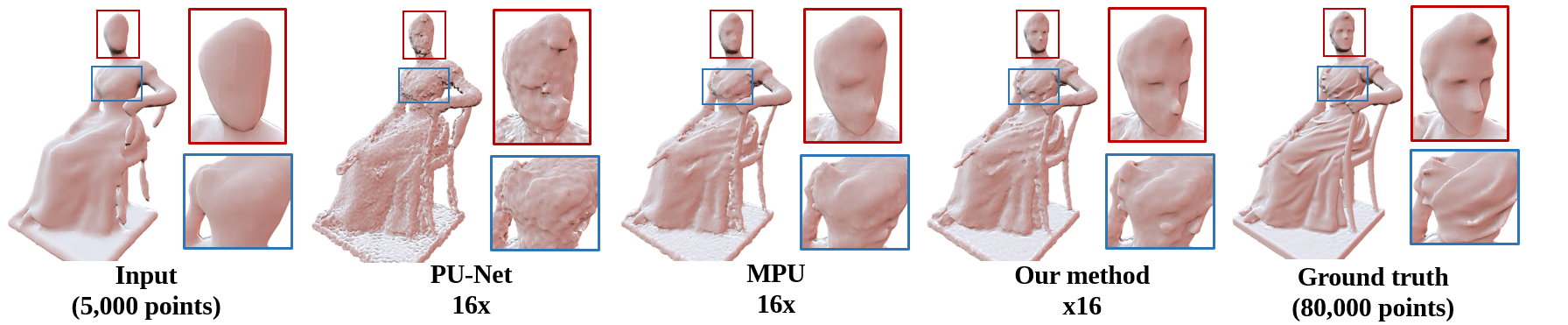}
     \end{subfigure} \vspace{-0.25in}
        \caption{\footnotesize{Visual comparisons for scanned 3D models. Each input sparse 3D point cloud has $M=5000$ points and upsampled by a factor $R=16$. We applied the same surface reconstruction algorithm to the sparse and densified points by different methods. For each data, the top and bottom rows correspond to the reconstructed surfaces and point clouds, respectively. As the close-up views, PUGeo-Net can handle the geometric details well. See the \textit{Supplementary file} for more visual comparisons and the video demo.}} 
        \label{fig:compare}
        \vspace{-0.3in}
\end{figure*}

\begin{figure*}[h!]
     \centering
     \includegraphics[width=0.8\textwidth]{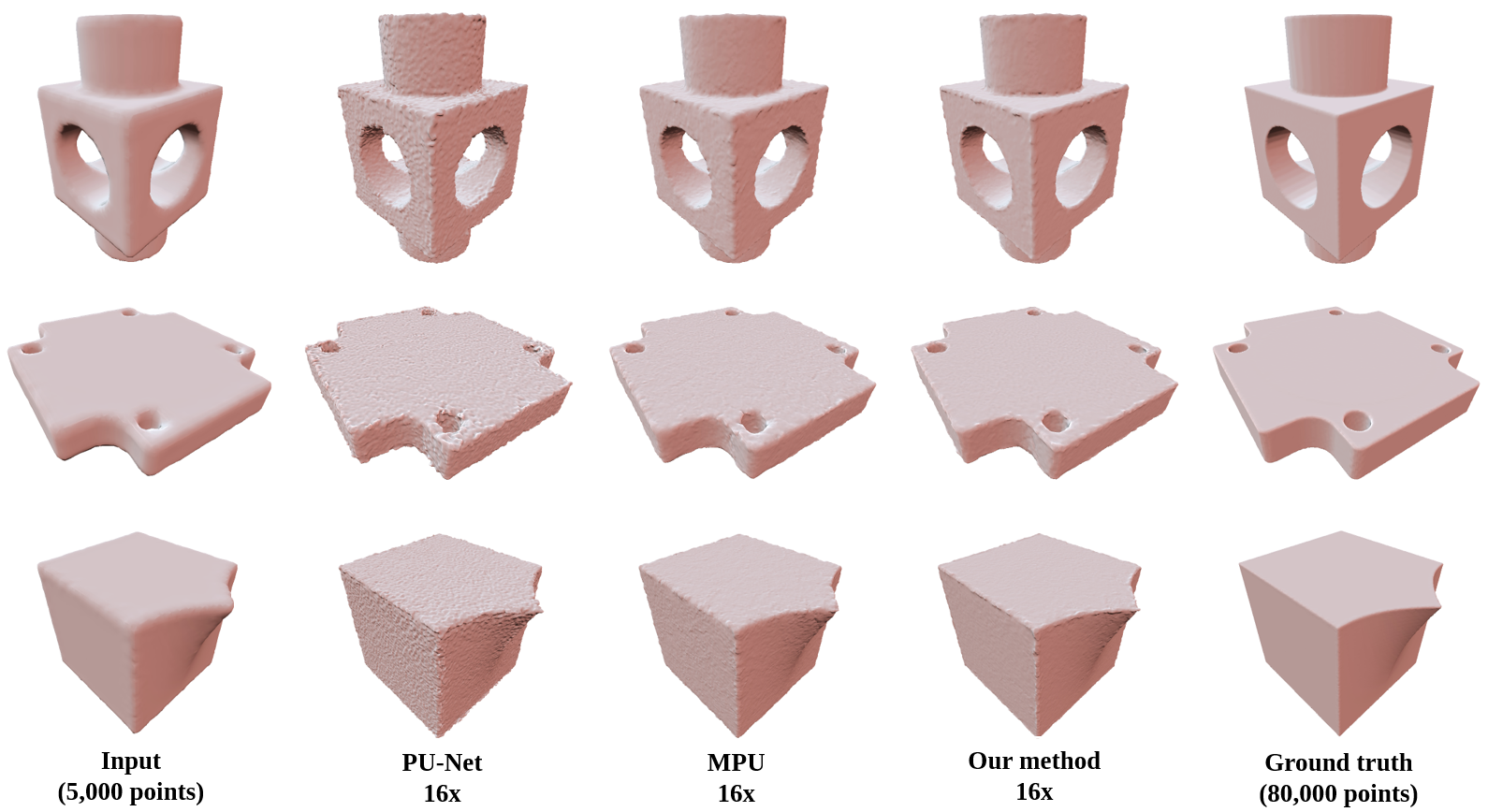}\vspace{-0.1in}
                  \caption{\footnotesize{Visual comparisons on CAD models with the upsampling factor $R=16$. The input point clouds have 5,000 points. We show the surfaces generated using the screened Poisson surface reconstruction (SPSR) algorithm \cite{kazhdan2013screened}. Due to the low-resolution of the input, SPSR fails to reconstruct the geometry. After upsampling, we observe that the geometric and topological features are well preserved in our results.}}
        \label{fig:toy}\vspace{-0.25in}
\end{figure*}

From Fig. \ref{fig:compare}, it can be observed that after performing upsampling the surfaces by PUGeo-Net present more geometric details and the best geometry structures, especially for the highly detailed parts with complex geometry, and they are closest to the groundtruth ones.
We also evaluated different methods on some man-made toy models. Compared with complex statue models, these man-made models consist of flat surfaces and sharp edges, which require high quality normals for surface reconstruction. As illustrated in Fig. \ref{fig:toy}, owing to the accurate normal estimation, PUGeo-Net can preserve the flatness and sharpness of the surfaces better than PU-Net and MPU. 
We further investigated how the quality of the reconstructed surface by PUGeo-Net changes with the upsample factor increasing. In Fig. \ref{fig:multirate}, it can be seen that as the upsample factor increases, PUGeo-Net can generate more uniformly distributed points, and the reconstructed surface is able to recover more details gradually to approach the groundtruth surface. \textit{See the supplementary material for more visual results and the video demo}.

\begin{figure}[h]
\centering
\vspace{-0.3in}
\includegraphics[width=0.7\textwidth]{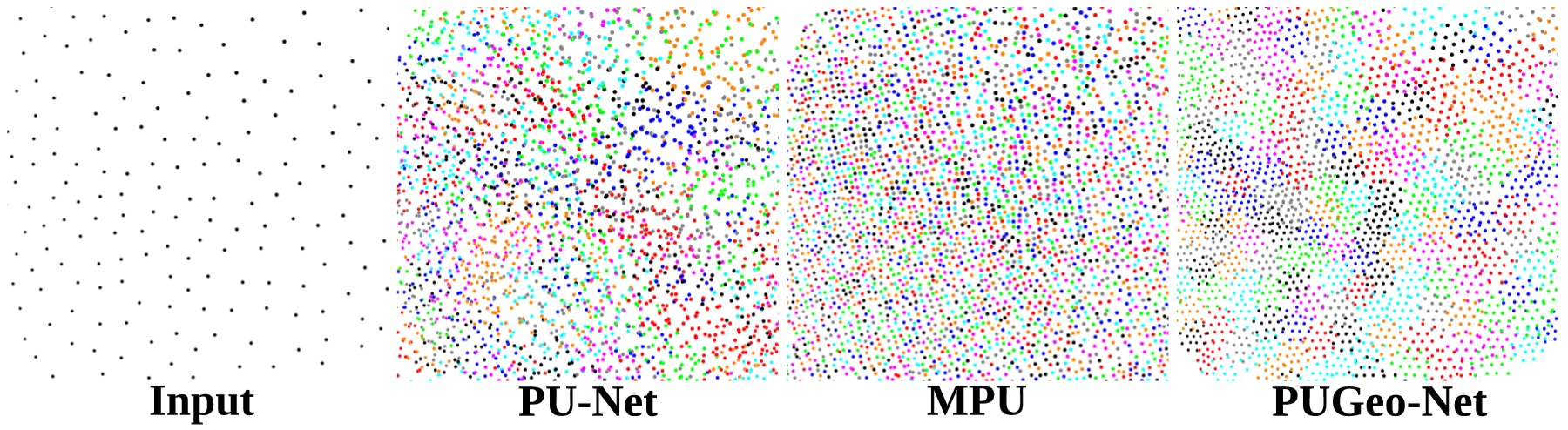}\vspace{-0.1in}
\caption{\footnotesize{Visual comparison of the distribution of generated 2D points with upsampling factor $R=16$. We distinguish the generated points by their source, assigned with colors. The points generated by PUGeo-Net are more uniform than those of PU-Net and MPU.} 
}
\label{fig:deform}
\end{figure}

\textbf{Comparison of the distribution of generated points.} In Fig. \ref{fig:deform}, we visualized a point cloud patch which was upsampled with 16 times by different methods. 
As PUGeo-Net captures the local structure of a point cloud elegantly in a geometry-centric manner, such that the upsampled points are uniformly distributed in the form of clusters. Using PUGeo-Net, the points generated from the same source point $\mathbf {x}_i$ are uniformly distributed in the local neighborhood $\mathbf{x}_i$, which justifies our parameterization-based sampling strategy.  PU-Net and MPU do not have such a feature. We also observe that our generated points are more uniform than theirs both locally and globally. 

     \vspace{-0.5in}
\subsection{Effectiveness of Normal Prediction}
\vspace{-0.1in}
Moreover, we also modified PU-Net, denoted as PU-Net-M, to predict coordinates and normals joinly by changing the neuron number of the last layer to 6 from 3. PU-Net-M was trained with the same training dataset as ours. 

\begin{table}[t]
\centering
\caption{\small{Verification of the effectiveness of our normal prediction. Here, the upsamplin raito $R$ is $8$. PCA-* indicates the normal prediction by PCA with various numbers of neighborhoods.}}
\scriptsize
\begin{tabular}{c||c|c|c||c|c|c|c}\Xhline{5\arrayrulewidth}
 \makebox[5em]{Methods} & \makebox[4em]{CD$^\#$}&\makebox[4em]{HD$^\#$}&\makebox[4em]{JSD$^\#$}&  \makebox[5em]{Methods} & \makebox[4em]{CD$^\#$}&\makebox[4em]{HD$^\#$}&\makebox[4em]{JSD$^\#$}  \\\Xhline{2\arrayrulewidth}
PCA-10 & 0.586 & 5.837 & 3.903 &PCA-15 & 0.577 & 5.893 & 3.789\\
PCA-25 & 0.575 & 5.823 & 3.668 &PCA-35 & 0.553 & 5.457 & 3.502\\
PCA-45 & 0.568 & 5.746 & 3.673 &PU-Net-M & 0.678 & 6.002 & 4.139\\
\Xhline{2\arrayrulewidth}
PUGeo-Net & \bf{0.549} & \bf{5.232} & \bf{3.464} \\\Xhline{5\arrayrulewidth}
\end{tabular}
\label{table:normal}\vspace{-0.5cm}
\end{table}

The quantitative results are shown in Table \ref{table:normal}, where we can see that (1) the surfaces reconstructed with the normals by PUGeo-Net produces the smallest errors for all the three metrics; (2) the number of neighborhoods in PCA based normal prediction is a heuristic parameter and influences the final surface quality seriously; and (3) the PU-Net-M achieves the worst performance, indicating that a naive design without considering the geometry characteristics does not make sense. 
\vspace{-0.15in}
\subsection{Robustness Analysis}
\vspace{-0.1in}
We also evaluated PUGeo-Net with non-uniform, noisy and real scanned data to demonstrate its robustness.\vspace{-0.2in}
\begin{figure}[h!]
\centering
\includegraphics[width=1\textwidth]{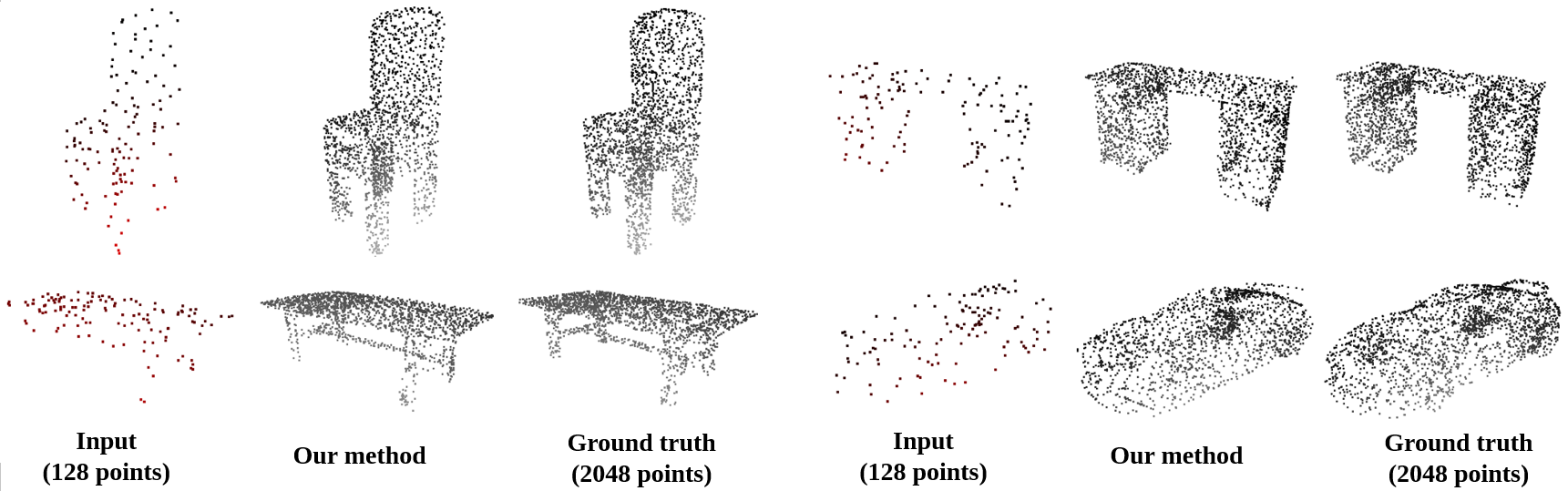}
\vspace{-0.25in}
\caption{16$\times$ upsampling results on non-uniformly distributed point clouds.}
\label{fig:nonuniform}\vspace{-0.2in}
\end{figure}

\vspace{-0.1in}
\textbf{Non-uniform data}. As illustrated in Fig. \ref{fig:nonuniform}, the data from ShapeNet \cite{wu20153d} were adopted for evaluation, where 128 points of each point cloud were randomly sampled without the guarantee of the uniformity. Here we took the upsampling task $R=16$ as an example. 
From Fig. \ref{fig:nonuniform}, it can be observed that PUGeo-Net can successfully upsample such non-uniform data to dense point clouds which are very close to the ground truth ones, such that the robustness of PUGeo-Net against non-uniformity is validated.

\begin{figure}[h!]
\centering
\includegraphics[width=1\textwidth]{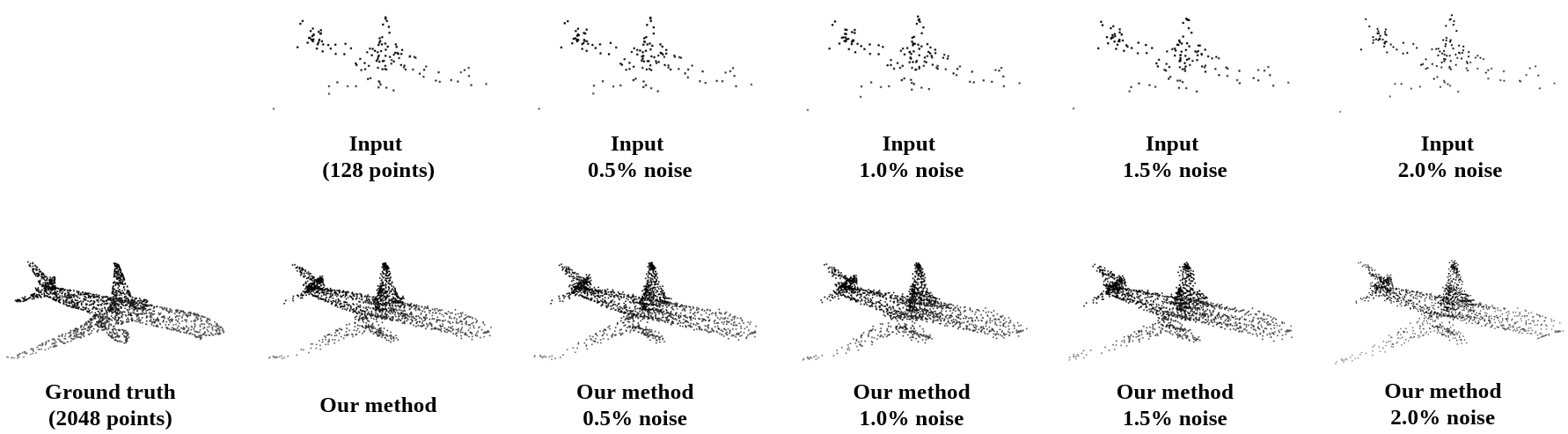}
\vspace{-0.25in}
\caption{\footnotesize{16$\times$ upsampling results on non-uniform point clouds with various levels of Gaussian noise.} }
\label{fig:noise}\vspace{-0.3in}
\end{figure}

\textbf{Noisy data}. 
We further added Gaussian noise to the non-uniformly distributed point clouds from ShapeNet, leading to a challenging application scene for evaluation, and 
various noise levels were tested. 
From Fig. \ref{fig:noise}, we can observe our proposed algorithm still works very on such challenging data, convincingly validating its robustness against noise.

\begin{figure}[h!]
\centering
\vspace{-0.4in}
\includegraphics[width=1\textwidth]{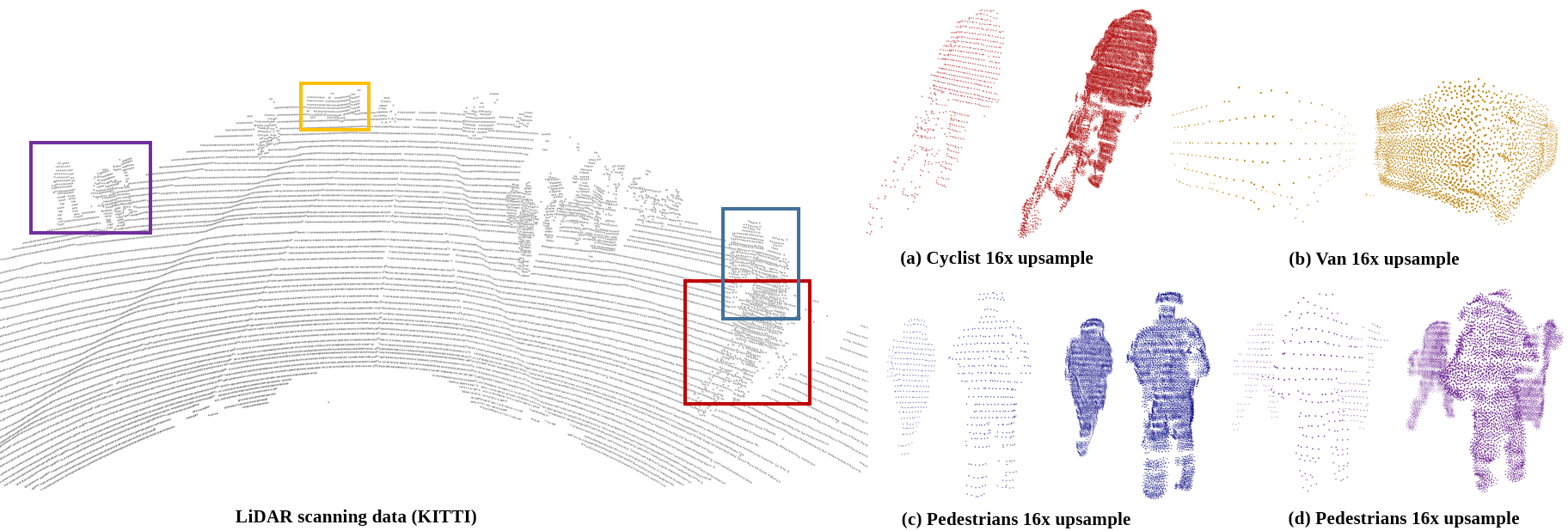}
\vspace{-0.25in}
\caption{\footnotesize{16$\times$ upsampling results on real scanned data by the LiDAR sensor.}}
\label{fig:scan}\vspace{-0.5cm}
\end{figure}

\textbf{Real scanned data}.  Finally, we evaluated PUGeo-Net with real scanned data by the LiDAR sensor \cite{Geiger2013IJRR}. Real scanned data contain noise, outliers, and occlusions. Moreover, the density of real scanned point clouds varies with the distance between the object and the sensor. As shown in Fig. \ref{fig:scan}, we can see our PUGeo-Net can produce dense point clouds with richer geometric details. 
\vspace{-0.2in}

\subsection{Ablation Study}
We conducted an ablation study towards our model to evaluate the contribution and effectiveness of each module. Table \ref{table:ablation} shows the quantitative results. Here we took the task with $R=8$ as an example, and similar results can be observed for other upsampling factors. 

\vspace{-0.35in}
\begin{table*}[h!]
\centering
\caption{\footnotesize{Ablation study. \textbf{Feature recalibration}: concatenate multiscale feature directly without the recalibration module. \textbf{Normal prediction}: only regress coordinates of points without normal prediction and supervision. \textbf{Learned adaptive 2D sampling}: use a predefined 2D regular grid as the parametric domain instead of the learned adaptive 2D smapling. \textbf{Linear transformation}: regress coordinates and normals by non-linear MLPs directly without prediction of the linear transformation. \textbf{Coarse to fine}: directly regress coordinates and normals without the intermediate coarse prediction.}}
\scriptsize
\begin{tabular}{r||c c c c c|c c c}\Xhline{5\arrayrulewidth}
 Networks  & \makebox[3em]{CD} & \makebox[3em]{HD}  &\makebox[3em]{JSD} & \makebox[3.5em]{P2F mean} & \makebox[3.5em]{P2F std} &\makebox[3em]{CD$^\#$} & \makebox[3em]{HD$^\#$} & \makebox[3em]{JSD$^\#$} \\\Xhline{2\arrayrulewidth}
Feature recalibration & 0.325 & 1.016 & 0.371 & 0.725 & 0.802 & 0.542 & 5.654 & 3.425 \\
Normal prediction & 0.331 & 2.232 & 0.427 & 0.785 & 0.973 & 0.563 & 5.884 & 3.565 \\
Learned adaptive 2D sampling & 0.326 & 1.374 & 0.407 & 0.701 & 0.811 & 0.552 & 5.758 & 3.456 \\
Linear transformation & 0.394 & \bf{1.005} & 1.627 & 0.719 & \bf{0.720} & 1.855 & 11.479 & 9.841 \\
Coarse to fine & 0.330 & 1.087 & 0.431 & 0.746 & 0.748 & 0.534 & \bf{5.241} & 3.348 \\\Xhline{2\arrayrulewidth}
Full model & \bf{0.323} & 1.011 & \bf{0.357}& \bf{0.694} & 0.808 & \bf{0.524} & 5.267 & \bf{3.279} \\\Xhline{5\arrayrulewidth} 
\end{tabular}
\label{table:ablation}\vspace{-0.3in}
\end{table*}

 From Table \ref{table:ablation}, we can conclude that (1) directly regressing the coordinates and normals of points by simply using MLPs instead of the linear transformation decreases the upsampling performance significantly, demonstrating the superiority of our geometry-centric design; 
(2) the joint regression of normals and coordinates are better than that of only coordinates; 
and (3) the other novel modules, including feature recalibration, adaptive 2D sampling, and the coarse to fine manner, all contribute to the final performance.

\begin{figure*}[h!]
     \centering
     \begin{subfigure}[b]{0.48\textwidth}
         \centering
         \includegraphics[width=\textwidth]{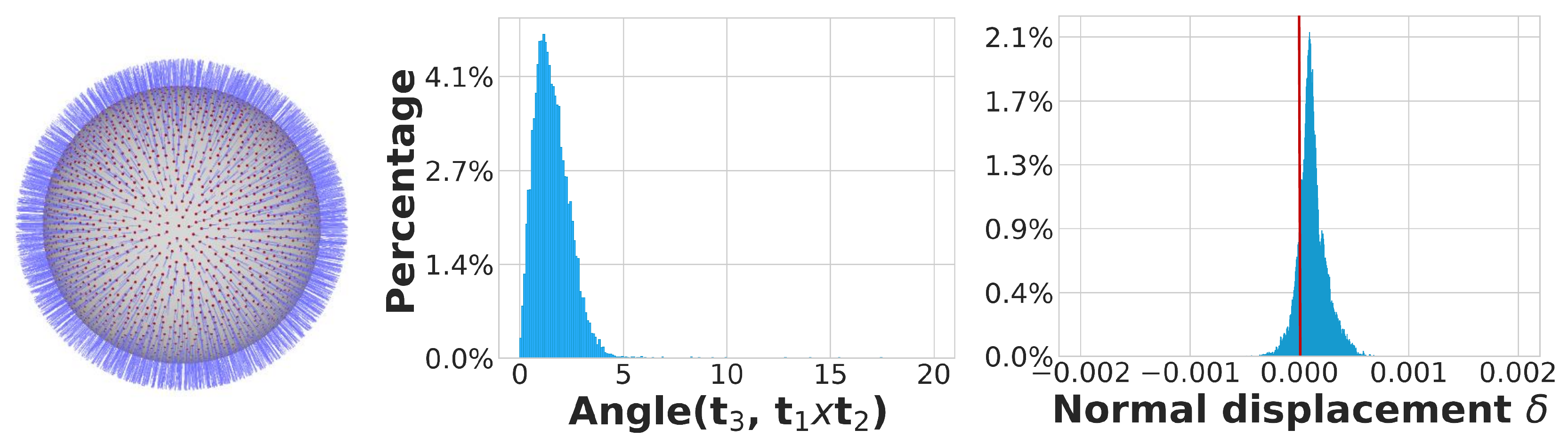}
     \end{subfigure}
     \begin{subfigure}[b]{0.48\textwidth}
         \centering
         \includegraphics[width=\textwidth]{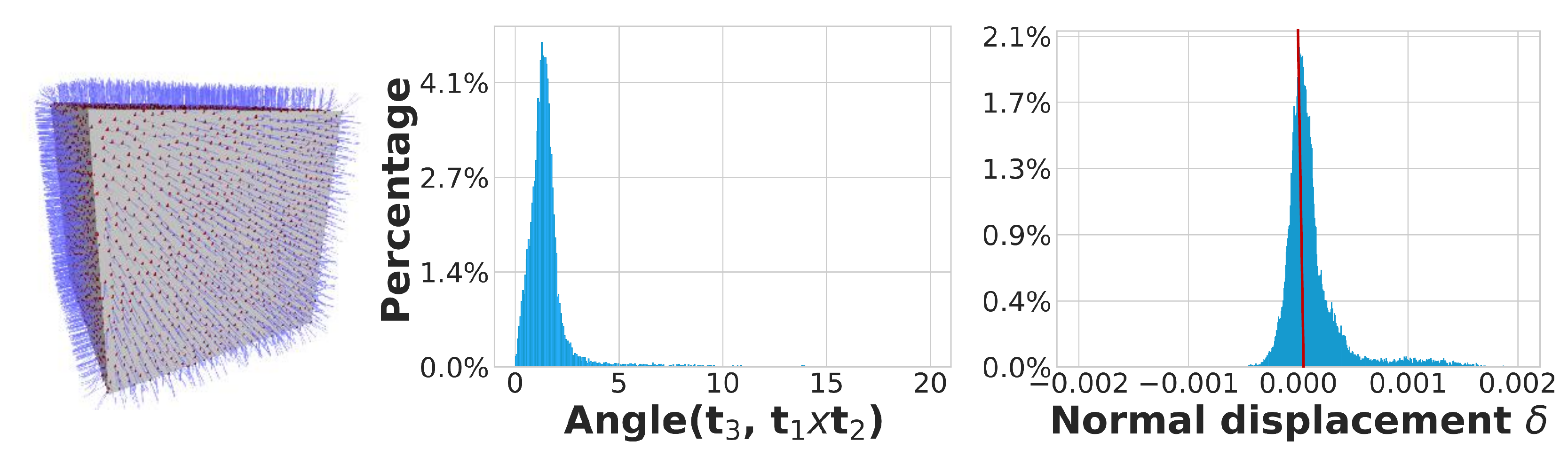}
     \end{subfigure}\vspace{-0.1in}
               \caption{\footnotesize{Statistical analysis of the predicted transformation matrix $\mathbf{T}=[\mathbf{t}_1; \mathbf{t}_2; \mathbf{t}_3]\in \mathbb{R}^{3\times 3}$ and normal displacement $\delta$, which can be used to fully reconstruct the local geometry.}}
        \label{fig:dist} \vspace{-0.3in}
\end{figure*}

To demonstrate the geometric-centric nature of PUGeo-Net, we examined the accuracy of the linear matrix $\mathbf{T}$ and the normal displacement $\delta$ for a unit sphere and a unit cube, where the ground-truths are available. We use angle $\theta$ to measure the difference of vectors $\mathbf{t}_3$ and $\mathbf{t}_1\times\mathbf{t}_2$, where $\mathbf{t}_i\in\mathbb{R}^{1\times 3}$ ($i=1,2,3$) is the $i$-th column of $\mathbf{T}$.  As Fig. \ref{fig:dist} shows, the angle $\theta$ is small with the majority less than 3 degrees, indicating high similarity between the predicted matrix $\mathbf{T}$ and the analytic Jacobian matrix. For the unit sphere model, we observe that the normal displacements $\delta$ spread in a narrow range, since the local neighborhood of $\mathbf{x}_i$ is small and the projected distance from a neighbor to the tangent plane of $\mathbf{x}_i$ is small. For the unit cube model, the majority of the displacements are close to zero, since most of the points lie on the faces of the cube which coincide with their tangent planes. On the other hand, $\delta$s spread in a relatively wide range due to the points on the sharp edges, which produce large normal displacement. 
\vspace{-0.2in}

\vspace{-0.2in}
\section{Conclusion and Future Work}
\vspace{-0.25cm}
We presented PUGeo-Net, a novel deep learning based framework for 3D point cloud upsampling. As the first deep neural network constructed in a geometry centric manner, PUGeo-Net has 3 features that distinguish itself from the other methods which are largely motivated by image super-resolution techniques. First, PUGeo-Net explicitly learns the first and second fundamental forms to fully recover the local geometry unique up to rigid motion; second, it adaptively generates new samples (also learned from data) and can preserve sharp features and geometric details well; third, as a by-product, it can compute normals of the input points and generated new samples, which make it an ideal pre-processing tool for the existing surface reconstruction algorithms. Extensive evaluation shows PUGeo-Net outperforms the state-of-the-art deep learning methods for 3D point cloud upsampling in terms of accuracy and efficiency.

PUGeo-Net not only brings new perspectives to the well-studied problem, but also links discrete differential geometry and deep learning in a more elegant way. In the near future, we will apply PUGeo-Net to more challenging application scenarios (e.g., incomplete dataset) and develop an end-to-end network for surface reconstruction. Since PUGeo-Net explicitly learns the local geometry via the first and second fundamental forms, we believe it has the potential for a wide range 3D processing tasks that require local geometry computation and analysis, including feature-preserving simplification, denoising, and compression. 
\clearpage

\end{document}